\documentclass[11pt]{article}
\usepackage{rotating}
\usepackage{acl2015}
\usepackage{times}
\usepackage{url}
\usepackage{latexsym}
\usepackage{tabularx}
\usepackage{makecell}
\usepackage{hyperref}
\usepackage{multirow}
\usepackage{multicol}
\usepackage{booktabs}
\usepackage{appendix}
\usepackage{CJKutf8}
\usepackage[section]{placeins}
\usepackage[hang,flushmargin]{footmisc} 
\usepackage[T1]{fontenc}
\usepackage{lipsum}
\newcommand\blfootnote[1]{%
  \begingroup
  \renewcommand\thefootnote{}\footnote{#1}%
  \addtocounter{footnote}{-1}%
  \endgroup
}
\usepackage{fancyhdr}
\usepackage{caption}
\usepackage{amsmath}
\usepackage{pifont}
\usepackage[table,xcdraw]{xcolor}
\usepackage{graphicx}

\title{DUMA: a Dual-Mind Conversational Agent with Fast and Slow Thinking}

\author{Xiaoyu Tian, Liangyu Chen, Na Liu, Yaxuan Liu,\\
\textbf{Wei Zou, Kaijiang Chen, Ming Cui}\\
Beike Inc., Beijing, China  \\
\texttt{\{tianxiaoyu011, chenliangyu003, liuna013, liuyaxuan002,} \\
\texttt{zouwei026, chenkaijiang001, cuiming001\}@ke.com}}

\begin{document}
\maketitle

\begin{abstract}
Inspired by the dual-process theory of human cognition, we introduce DUMA, a novel conversational agent framework that embodies a dual-mind mechanism through the utilization of two generative Large Language Models (LLMs) dedicated to fast and slow thinking respectively. The fast thinking model serves as the primary interface for external interactions and initial response generation, evaluating the necessity for engaging the slow thinking model based on the complexity of the complete response. When invoked, the slow thinking model takes over the conversation, engaging in meticulous planning, reasoning, and tool utilization to provide a well-analyzed response. This dual-mind configuration allows for a seamless transition between intuitive responses and deliberate problem-solving processes based on the situation. We have constructed a conversational agent to handle online inquiries in the real estate industry. The experåiment proves that our method balances effectiveness and efficiency, and has a significant improvement compared to the baseline.

\end{abstract}

\section{Introduction}
In the era of rapid progress of LLMs\cite{gpt3.5,gpt4}, creating conversational agents\cite{wang2023survey,xi2023rise,weng2023llmpowered} that can emulate human-like interactions is both a challenge and an aspiration. Drawing inspiration from the dual-process theory\cite{daniel2017thinking} of human cognition, which proposes two distinct cognitive processes—a fast, intuitive one and a slower, analytical one. we introduce DUMA, a novel conversational agent framework\cite{zalta1995stanford}. While there have been efforts like SwiftSage\cite{lin2023swiftsage} that delve into integrating fast and slow thinking processes in AI agents, DUMA stands out by prioritizing conversational scenarios.

DUMA, symbolizing a Dual-Mind Conversational Agent, merges two generative Large Language Models (LLMs) to accommodate distinct cognitive processes: fast and slow thinking. The Fast Mind module of DUMA stands out for its agility and efficiency, readily addressing straightforward scenarios. However, its nimbleness might falter in more intricate situations. Conversely, the Slow Mind takes its time, operating at a deliberate pace which might seem less efficient. Yet, this deliberation equips it to grapple with complex challenges, particularly those demanding in-depth reasoning or the invocation of external tools. Together, these two minds allow DUMA to deliver a balanced conversational experience, oscillating smoothly between immediate replies and profound problem-solving.

DUMA's dual-mind structure is reminiscent of human cognitive processes, differentiating between the Fast Mind and the Slow Mind. For routine queries, the Fast Mind takes the lead with immediate responses. However, when complex queries arise that demand deeper analysis, like mathematical or logical challenges, it calls upon the Slow Mind. Unlike its counterpart, the Slow Mind doesn't interact directly with users. Instead, it delves into the problem using a methodology inspired by ReAct\cite{yao2023react}, often calling upon external tools for assistance. Once its comprehensive analysis is done, the insights are relayed back to the Fast Mind, which crafts the response. Importantly, these insights are archived in the Fast Mind's "Memory Area", ensuring efficiency in future related dialogues, epitomizing DUMA's blend of agility and depth.

We've conducted experiments in Chinese real estate online communication scenarios, confirming the efficacy of our approach. Although our experiments were specific to the Chinese context and real estate domain, we believe that the methodology behind DUMA possesses a broader applicability.

Our main contributions in this paper are two folds:

\begin{itemize}
\item Introducing DUMA, a novel conversational agent framework built upon the dual-process theory, integrating two LLMs for fast and slow cognitive processes.

\item Demonstrating the application and efficacy of DUMA in real-world scenarios, specifically in the real estate industry, where it showcases significant improvements over baseline models.
\end{itemize}

\section{Methodology}

In this section, we will first introduce the overall structure of DUMA, then set forth the design and operation process of Fast Mind and Slow Mind respectively, and finally elaborate the internal and external interaction of DUMA.

\subsection{DUMA Overall Structure}

\begin{figure*}[hbt]
\vspace{0.1cm}
\centering
\setlength{\abovecaptionskip}{0.1cm} 
\includegraphics[scale=0.33]{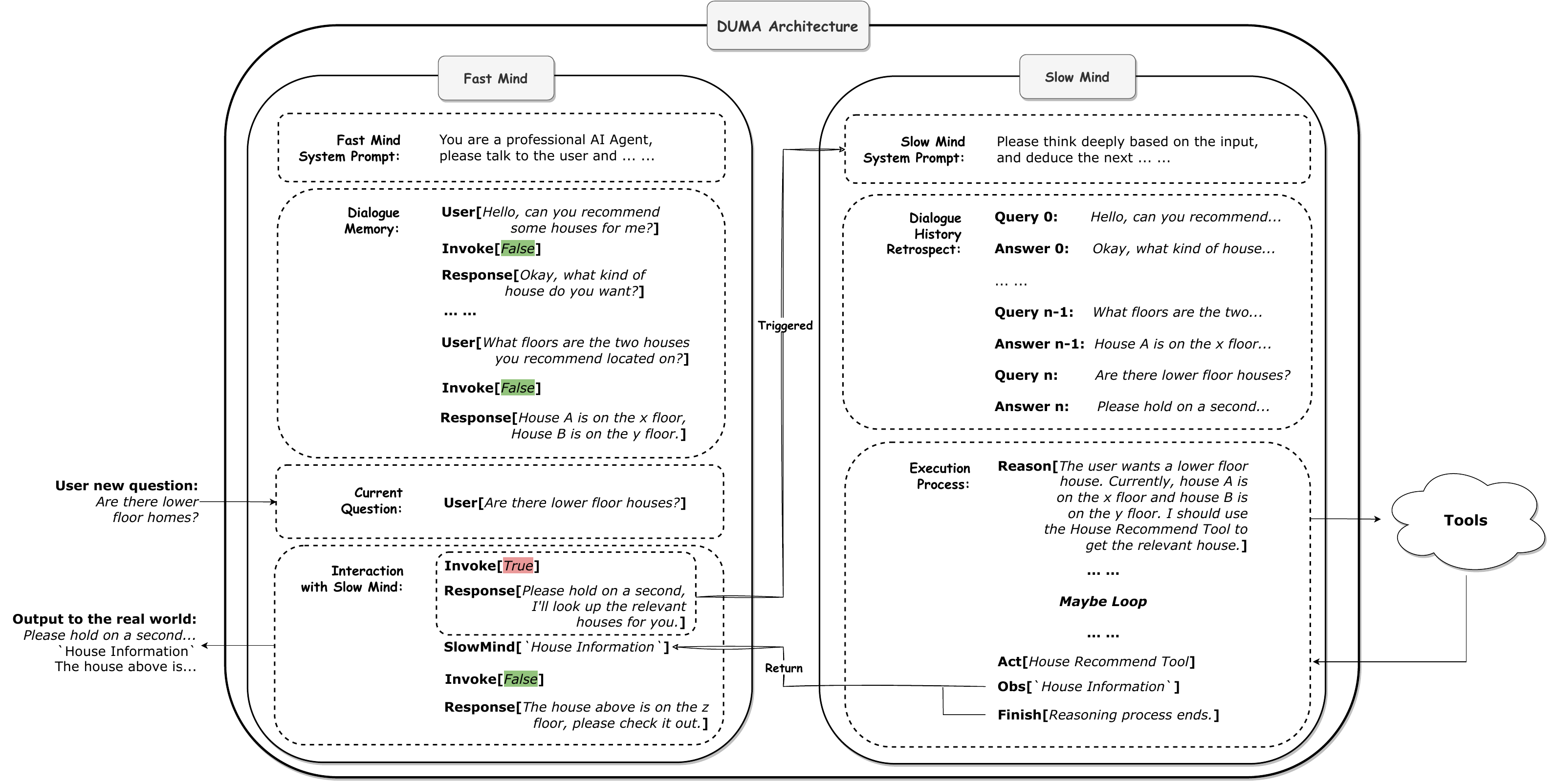}
\caption{Upon receiving a user question, both the question and Dialogue Memory are fed into Fast Mind. If deep analysis is deemed necessary, Slow Mind is activated. Slow Mind reviews historical dialogues, conducts reasoning, takes action, and observes. The results are relayed back to Fast Mind, which crafts a response based on Slow Mind's feedback, delivering it to the real-world context.}
\label{duma_structure}
\end{figure*}

The thinking center of DUMA contains two minds, Fast Mind (defined as $Mind_{Fast}$) and Slow Mind (defined as $Mind_{Slow}$), as shown in the Figure \ref{duma_structure}.

As discussed above, simple questions in general conversations do not require in-depth thinking. $Mind_{Fast}$ is responsible for quickly thinking, replying to those simple questions, and directly interacting with human. For those difficult questions, such as mathematics, complex reasoning, etc., after $Mind_{Fast}$ perceives the complexity of the question, it will send a signal to $Mind_{Slow}$, thereby triggering DUMA's deep thinking mechanism.

$Mind_{Slow}$ will not directly interact with human. After deep thinking, $Mind_{Slow}$ will transmit the results and data obtained after calling the tool back to $Mind_{Fast}$. The current turn of dialogue will be generated by $Mind_{Fast}$ based on the inference results of $Mind_{Slow}$. Finally, DUMA will return the result through $Mind_{Fast}$.

\subsection{Fast Mind}

For each turn of dialogue, $Mind_{Fast}$ may have two input sources, human utterance or $Mind_{Slow}$ thinking results. $Mind_{Fast}$ infers each turn of responses based on historical dialogue and current input.

Formally, for time step $t$, assuming the human utterance is $Q_t$, the $Mind_{Slow}$ thinking result is $S_t$, the input will be processed into the following format:

\begin{align}
I^{Fast}_t = \begin{cases} User[Q_t], & Human \ Input \\ SlowMind[S_t], & Mind_{Slow} \ Input\end{cases}
\label{I_t_Fast}
\end{align}

According to the previous dialogue, $Mind_{Fast}$ quickly thinking and generating answer. The results generated by $Mind_{Fast}$ can indicate whether $Mind_{Slow}$ needs to be triggered and what needs to be replied. $Mind_{Fast}$'s response in round $t$ is defined as $O_t$:

\begin{align}
O^{Fast}_t = 
\begin{aligned}
&Invoke[V], V \in \{True, False\} \\
&Response[Mind_{Fast} \ output_t]
\end{aligned}
\end{align}

When the value in "Invoke" is True, $Mind_{Slow}$ will be awakened. At this time, DUMA will think deeply and thoroughly. The content in "Response" is the reply to the t-th turn of conversation.

Analogous to human, $Mind_{Fast}$ conducts multi-turn of dialogue with individuals. For the current turn of response, $Mind_{Fast}$ needs to "consider" the context of past dialogue, therefore, we need to process historical conversation into multi-turn format that the model can understand:

\small
\begin{align}
\begin{aligned}
    &Context_t = \\
    &\underbrace{M_bI_0M_eO_0}_{conv_0}\underbrace{M_bI_1M_eI_1}_{conv_1} \dots \underbrace{M_bI_{t-1}M_eO_{t-1}}_{conv_{t-1}}M_bI_tM_e
\end{aligned}
\end{align}
\normalsize

$M_b$ and $M_e$ are the pattern of the LLM's multi-turn dialogue\textsuperscript{1}\blfootnote{
    \textsuperscript{1}The pattern of multi-turn of dialogue may be different for different open source LLMs.
}. $Mind_{Fast}$ performs quick thinking based on $Context_t$, and generates the result $O_t$:

\begin{align}
    O^{Fast}_t = Mind_{Fast}(Context_t)
\end{align}

\subsection{Slow Mind}
\label{slow_mind}

When $Mind_{Slow}$ is awakened, just like human, $Mind_{Slow}$ will first retrospect the past dialogue. Assume that the $Mind_{Fast}$ response in round $t$ is $A_t$, and the past dialogue will be structured by $Mind_{Fast}$ in the following format:

\begin{align}
Dialogue^{his}_{t} = \left\{
\begin{aligned}
& Query \ 0: Q_0 \\
& Answer \ 0: A_0 \\
& \cdots \\
& Query \ t-1: Q_{t-1} \\
& Answer \ t-1: A_{t-1} \\
& Query \ t: Q_t \\
\end{aligned}
\right.
\end{align}

Upon encapsulation of $Dialogue^{his}_{t}$ by $Mind_{Fast}$, the content is subsequently conveyed to $Mind_{Slow}$ to denote the historical dialogue exchanged between $Mind_{Fast}$ and individuals. Subsequent to this, $Mind_{Slow}$ engages in reasoning, action, and observation:

\begin{align}
O^{Slow}_t = Mind_{Slow}(Dialogue^{his}_{t})
\end{align}

Inspired by ReAct\cite{yao2023react}, we divide the $Mind_{Slow}$ thinking process into four parts (Reason, Act, Obs, Finish), as shown in the Figure \ref{duma_structure}.

When encountering complex problems, $Mind_{Slow}$ may reason in multiple steps. For example, when external tool support is needed, $Mind_{Slow}$ calls the tool through "Act" and observes the results of the tool through "Obs". Then the current review of the t-th turn of dialogue, all past reasoning, action and observation will be encapsulated into a new chain of thought and input into $Mind_{Slow}$ which will judge whether the next step of in-depth thinking is needed. If necessary, the process stated above will be looped, otherwise will be finished.

The value in "Obs" or "Finish" is the final reasoning result of $Mind_{Slow}$, which is encapsulated into the Formula \ref{I_t_Fast} ($Mind_{Slow} \ Input$) and then input into $Mind_{Fast}$.

\begin{figure*}[hbt]
\vspace{0.1cm}
\centering
\setlength{\abovecaptionskip}{0.1cm} 
\includegraphics[scale=0.55]{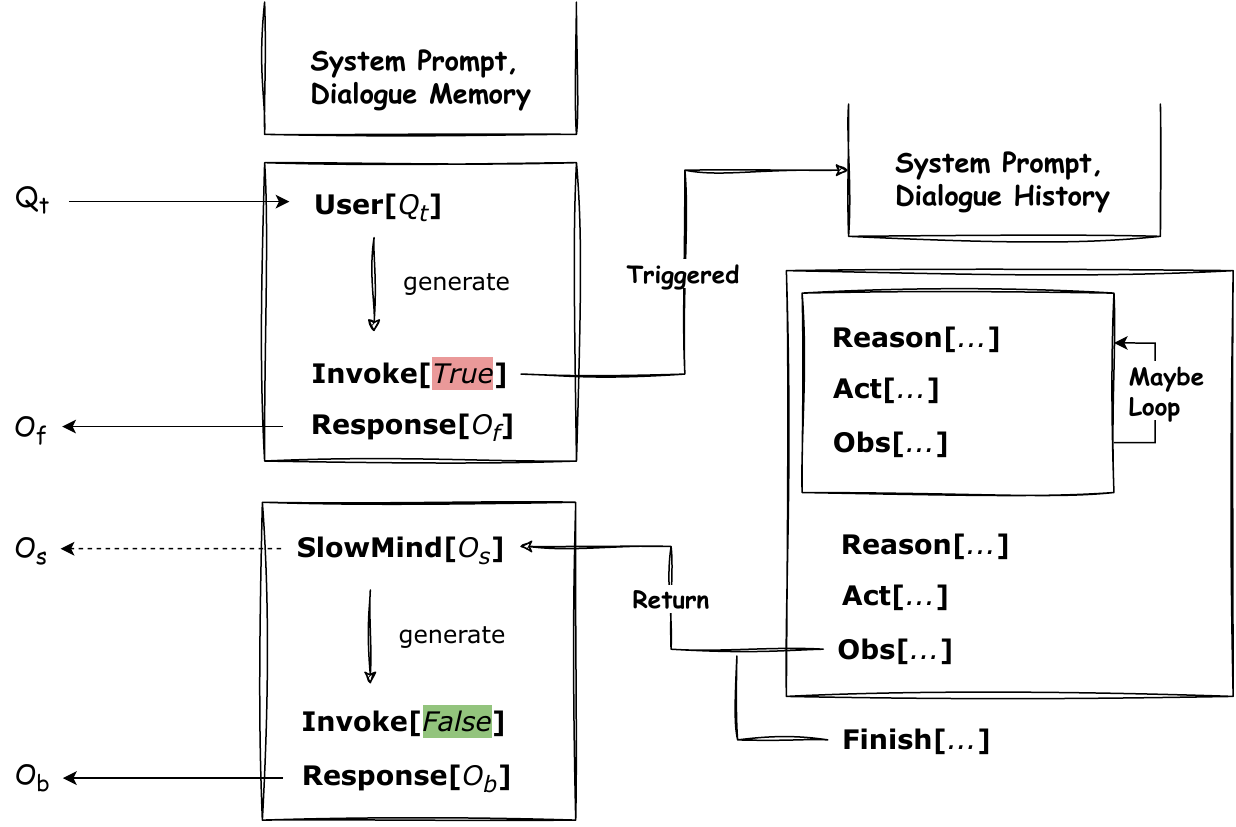}
\caption{Fast Mind produces the current result based on Dialogue Memory and $Q_t$. When Invoke is set to True, the deep thinking mechanism of Slow Mind is triggered. Slow Mind then returns its observation and reasoning results to Fast Mind. Whether $O_s$ is relayed to the real world is determined by the system. Ultimately, $O_f$, $O_s$, and $O_b$ are returned to the real world as the complete results of the current round.}
\label{duma_interact}
\end{figure*}

\subsection{DUMA: Fast Mind + Slow Mind}

$Mind_{Fast}$ interacts with the real world to obtain the latest question and interacts with the $Mind_{Slow}$ (if necessary) to obtain in-depth thinking results. "Dialogue Memory" shown in Figure \ref{duma_structure} consists of two parts: the interaction between $Mind_{Fast}$ and the real world, and the results returned by $Mind_{Slow}$ (including observation, inference results, etc.). For the same situation that needs to trigger $Mind_{Slow}$, DUMA only needs to think deeply once, and the results are saved in the "Dialogue Memory" of $Mind_{Fast}$ in the form of dialogue. When asked questions related to memory, $Mind_{Fast}$ quickly generates and responds based on the interactive memory with the real world and $Mind_{Slow}$, thereby improving the conversation efficiency of the entire agent.

As shown in the Figure \ref{duma_interact}, when DUMA receives a new question, the system prompt, historical data and current questions will be spliced and input into $Mind_{Fast}$ for generating. When $Mind_{Slow}$'s in-depth thinking is triggered, system prompt and dialogue reviews (described in Section \ref{slow_mind}) are spliced and input into $Mind_{Slow}$ for reasoning, actions, and observations. This process may be looped, and $Mind_{Slow}$ decides when to end on its own. After $Mind_{Slow}$'s thinking is completed, the information in "Obs" is returned to $Mind_{Fast}$. $Mind_{Fast}$ makes inference responses based on the results of $Mind_{Slow}$, and finally returns $O_f$ and $O_b$ to the real world. The specific application decides whether $O_s$ needs to be exposed.

\section{Experiments}

We can employ LLMs to construct an agent grounded on the DUMA architecture without training. However, given the complexity of online real estate conversation scenarios, Fast Mind employs Baichuan-13B-Chat\cite{Baichuan13Bgithub} as its foundational model for training. Simultaneously, considering that Slow Mind might necessitate multi-step reasoning, ChatGLM2-6B\cite{zeng2022glm,du2022glm} is utilized as the base model for training to enhance the agent's performance and expedite the inference efficiency.

\subsection{Data Collection}

\begin{figure*}[hbt]
\vspace{0.1cm}
\centering
\setlength{\abovecaptionskip}{0.1cm} 
\includegraphics[scale=0.6]{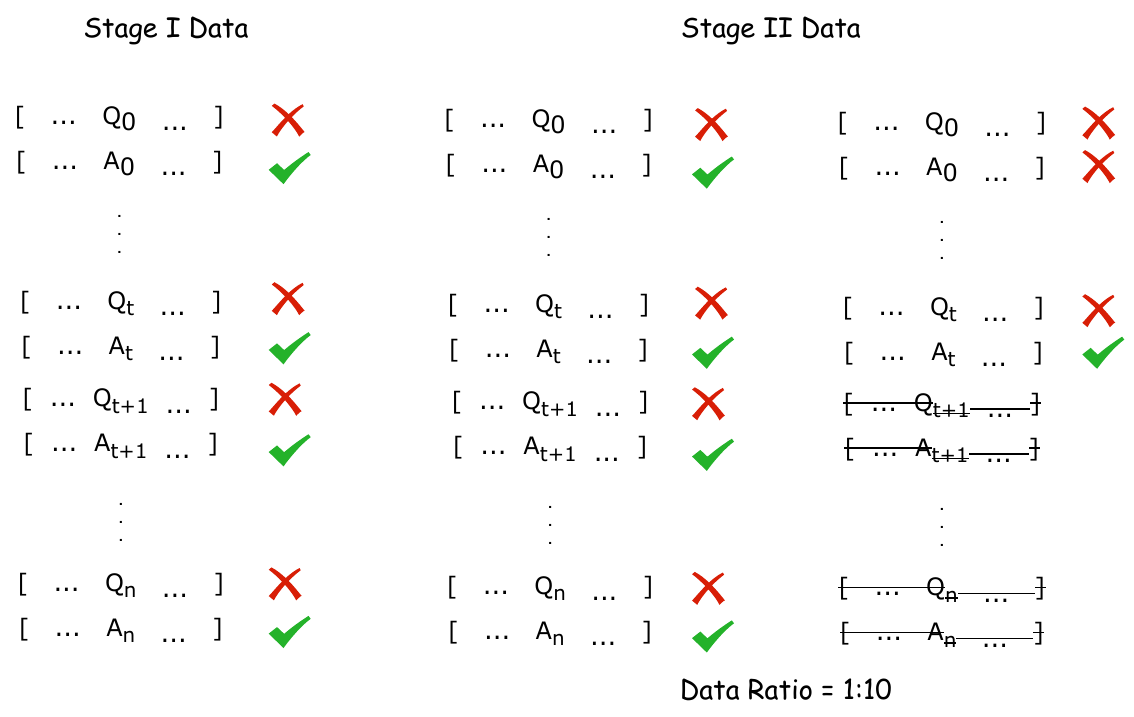}
\caption{Differences in data construction and loss computation process between SFT Stage I (Dialogue Training) and Stage II (Factuality Enhancement). In Stage II, if factuality calibration is performed for $A_t$, subsequent dialogues are deleted to ensure dialogue coherence. In the Figure, "\ding{51}" indicates that the loss needs to be calculated, while "\ding{55}" indicates that the loss doesn't need to be calculated. For both Stage I and Stage II, loss is computed for each turn of the complete dialogue, while for Stage II's factuality calibration data, only the annotated replies require loss computation.}
\label{duma_data_construct2}
\end{figure*}

\subsubsection{Fast Mind}
\label{fast_mind_data_collection}

Upon training with online real estate dialogue datasets, the model assimilates the conversational style and tone characteristic of actual individuals. However, we concurrently identify a concerning issue related to hallucination, such as consistency discrepancies, notably when the Fast Mind assimilates inputs from the Slow Mind, leading to the generation of erroneous content, as well as fabricating facts problems. To enhance the model's factuality, we implemented a two-phase SFT process: Dialogue Training followed by Factuality Enhancement.

\textbf{SFT Stage I - Dialogue Training} \quad We obtain original 13M dialogue data from online dialogue logs and established a data processing pipeline, which mainly includes data desensitization, rule based filtering, logic optimization, intent balancing, etc. Finally, we collect 66349 multi-turn dialogues. Before training, the data will be processed into the model's multi-turn dialogue format. The first stage SFT data uses complete dialogue.

\textbf{SFT Stage II - Factuality Enhancement} \quad To enhance the factuality of the model, we resampled 4M dialogues from online logs. After pipeline data processing (in order to ensure that the intent distribution is consistent with online, intent balancing is not performed), we acquire 65009 multi-turn dialogues. Based on the online probability distribution of intent and the likelihood of a given intent appearing at various positions within a dialogue, we randomly sampled 3,000 dialogues for factual calibration annotation. During annotation, only the response to the current turn of question required factual calibration. Given that the annotation might alter the logical coherence of subsequent dialogue, to ensure logical coherence, any content following the annotated dialogue is removed. The differences in data construction methods between Stage I and Stage II are illustrated in the Figure \ref{duma_data_construct2}.

\begin{table*}[t!]
\caption{The evaluation criteria details of the defined metrics.}
\label{metric_cri}
\centering
\scalebox{0.75}{
\begin{tabular}{lllllll}
\hline
{\color[HTML]{000000} \textbf{Score}} & {\color[HTML]{000000} \textbf{\begin{tabular}[c]{@{}l@{}}House\\ Expertise\end{tabular}}}                                               & {\color[HTML]{000000} \textbf{\begin{tabular}[c]{@{}l@{}}Tool\\ Calling Ability\end{tabular}}}                                        & {\color[HTML]{000000} \textbf{\begin{tabular}[c]{@{}l@{}}Industry\\ Familiarity\end{tabular}}}                                          & {\color[HTML]{000000} \textbf{\begin{tabular}[c]{@{}l@{}}Service\\ Attitude\end{tabular}}}                                                                        & {\color[HTML]{000000} \textbf{\begin{tabular}[c]{@{}l@{}}Demand\\ Mining\end{tabular}}}                                                   & {\color[HTML]{000000} \textbf{\begin{tabular}[c]{@{}l@{}}Promote\\ Invitation\end{tabular}}}                                                        \\ \hline
{\color[HTML]{000000} 0}              & {\color[HTML]{000000} \begin{tabular}[c]{@{}l@{}}All answered\\ incorrectly.\end{tabular}}                                              & {\color[HTML]{000000} \begin{tabular}[c]{@{}l@{}}All used\\ incorrectly.\end{tabular}}                                                & {\color[HTML]{000000} \begin{tabular}[c]{@{}l@{}}All answered\\ incorrectly.\end{tabular}}                                              & {\color[HTML]{000000} \begin{tabular}[c]{@{}l@{}}Not like\\ human, or\\ logically\\ confusing.\end{tabular}}                                              & {\color[HTML]{000000} \begin{tabular}[c]{@{}l@{}}No mining\\ action was\\ performed when\\ needed, or all failed.\end{tabular}}           & {\color[HTML]{000000} \begin{tabular}[c]{@{}l@{}}No invitation\\ action was\\ performed when\\ needed, or\\ all failed.\end{tabular}}               \\ \hline
{\color[HTML]{000000} 1}              & {\color[HTML]{000000} \begin{tabular}[c]{@{}l@{}}Answer at least\\ half of the\\ questions\\ successfully.\end{tabular}}                & {\color[HTML]{000000} \begin{tabular}[c]{@{}l@{}}Call at least\\ half of the\\ tools successfully.\end{tabular}}                      & {\color[HTML]{000000} \begin{tabular}[c]{@{}l@{}}Answer at least\\ half of the\\ questions\\ successfully.\end{tabular}}                & {\color[HTML]{000000} \begin{tabular}[c]{@{}l@{}}Like human\\ and logically\\ correct.\end{tabular}}                                                      & {\color[HTML]{000000} \begin{tabular}[c]{@{}l@{}}Mining action\\ at the right time\\ and obtain the\\ demand at least once.\end{tabular}} & {\color[HTML]{000000} \begin{tabular}[c]{@{}l@{}}Make an invitation\\ at least once at the\\ right time and do\\ not invite randomly.\end{tabular}} \\ \hline
{\color[HTML]{000000} 2}              & {\color[HTML]{000000} \begin{tabular}[c]{@{}l@{}}All answered\\ correctly and\\ can proactively\\ extend the\\ questions.\end{tabular}} & {\color[HTML]{000000} \begin{tabular}[c]{@{}l@{}}All used\\ correctly and\\ can proactively\\ use the relevant\\ tools.\end{tabular}} & {\color[HTML]{000000} \begin{tabular}[c]{@{}l@{}}All answered\\ correctly and\\ can proactively\\ extend the\\ questions.\end{tabular}} & {\color[HTML]{000000} \begin{tabular}[c]{@{}l@{}}Have strong\\ affinity, actively\\ respond to\\ questions and\\ proactively look\\ for new topics.\end{tabular}} & {\color[HTML]{000000} \begin{tabular}[c]{@{}l@{}}Mining action at\\ the right time\\ and obtain all\\ the demands.\end{tabular}}          & {\color[HTML]{000000} \begin{tabular}[c]{@{}l@{}}Make Invitation\\ action in all the\\ right time.\end{tabular}}                                    \\ \hline
\end{tabular}
}
\end{table*}

\begin{table*}[t!]
\caption{The evaluation results of the defined metrics.}
\label{results}
\centering
\begin{tabular}{lllllll}
\hline
{\color[HTML]{000000} \textbf{Model}}    & {\color[HTML]{000000} \textbf{\begin{tabular}[c]{@{}l@{}}House\\ Expertise\end{tabular}}} & {\color[HTML]{000000} \textbf{\begin{tabular}[c]{@{}l@{}}Tool Calling\\ Ability\end{tabular}}} & {\color[HTML]{000000} \textbf{\begin{tabular}[c]{@{}l@{}}Industry\\ Familiarity\end{tabular}}} & {\color[HTML]{000000} \textbf{\begin{tabular}[c]{@{}l@{}}Service\\ Attitude\end{tabular}}} & {\color[HTML]{000000} \textbf{\begin{tabular}[c]{@{}l@{}}Demand\\ Mining\end{tabular}}} & {\color[HTML]{000000} \textbf{\begin{tabular}[c]{@{}l@{}}Promote\\ Invitation\end{tabular}}} \\ \hline
{\color[HTML]{000000} ChatGPT$_{react}$} & {\color[HTML]{000000} 0.733}                                                              & {\color[HTML]{000000} 0.615}                                                                   & {\color[HTML]{000000} 0.769}                                                                   & {\color[HTML]{000000} 1.067}                                                               & {\color[HTML]{000000} 0.800}                                                            & {\color[HTML]{000000} 0.667}                                                                 \\ \hline
{\color[HTML]{000000} DUMA$_{StageI}$}   & {\color[HTML]{000000} 0.826}                                                              & {\color[HTML]{000000} \textbf{1.481}}                                                          & {\color[HTML]{000000} 1.091}                                                                   & {\color[HTML]{000000} 1.571}                                                               & {\color[HTML]{000000} \textbf{1.407}}                                                   & {\color[HTML]{000000} 1.214}                                                                 \\ \hline
{\color[HTML]{000000} DUMA}              & {\color[HTML]{000000} \textbf{1.550}}                                                     & {\color[HTML]{000000} 1.417}                                                                   & {\color[HTML]{000000} \textbf{1.125}}                                                          & {\color[HTML]{000000} \textbf{1.810}}                                                      & {\color[HTML]{000000} 1.357}                                                            & {\color[HTML]{000000} \textbf{1.471}}                                                        \\ \hline
\end{tabular}
\end{table*}

\subsubsection{Slow Mind}

We sample 0.5M raw dialogues from online logs. To facilitate better collaboration between Fast Mind and Slow Mind, the data underwent processing through a pipeline identical to Section \ref{fast_mind_data_collection} (excluding intent balancing), yielding 8,000 samples. To reduce the labeling cost, we label the data (has been desensitized) by GPT-4 pre-labeling and then manual correction. The method for data construction is elaborated in Section \ref{slow_mind}.

\subsection{Experiments Setups}

When training Fast Mind, the questions in the multi-turn dialogue of the first stage of SFT are not involved in the loss calculation. In the second stage of SFT, only the last turn of responses are involved in the loss calculation, and the gradients of the remaining questions and answers will be masked. To make the model better retain the dialogue logical coherence, during the second stage of training, we randomly sample 300 multi-turn dialogues from the first stage SFT training data (data ratio is 1:10) to conduct mixed training. The two-stage gradient calculation method is shown in the Figure \ref{duma_data_construct2}.

We use the ChatGPT3.5 interface and adopt the ReAct\cite{yao2023react} method as the baseline (denote as ChatGPT$_{react}$). When performing the first stage SFT of Fast Mind, we use Baichuan-13B-Chat as the basic model with learning rate of 1e-4, the second stage SFT uses the first stage checkpoint, and Slow Mind training with ChatGLM2-6B. The learning rate of the second stage SFT of Fast Mind and Slow Mind training are both 1e-5. Throughout all training procedures, the maximum length is 4096, training for 4 epochs, with a batch size of 32. We use a cosine LR schedule down to 10\% of the original learning rate, with 3\% warmup. All the models are trained with BFloat16 mixed precision for training stability.

\subsection{Metrics}

To effectively compare the capabilities of different system architectures, we conduct a manual assessment based on two primary dimensions: \textit{Knowledge} and \textit{Reasoning}.

\textit{Knowledge} competency is assessed in three areas: \textit{House Expertise}, \textit{Tool Calling Ability}, and \textit{Industry Familiarity}. \textit{House Expertise} gauges the agent's ability to respond to human utterance about housing, \textit{Tool Calling Ability} evaluates the capability of the agent's use of tools, and \textit{Industry Familiarity} measures the agent's general knowledge in the real estate domain.

\textit{Reasoning} ability encompasses three evaluation metrics: \textit{Service Attitude}, \textit{Demand Mining}, and \textit{Promote invitation}. \textit{Service Attitude} evaluates whether the agent interacts pleasantly and responds in a human-like manner. \textit{Demand Mining} assesses the agent's efficacy in uncovering and understanding the latent needs of the user during interactions. \textit{Promote Invitation} gauges the agent's aptitude in seeking contact details from humans and inviting them for offline meetings at appropriate times.

Referring to the intent distribution of online dialogue logs, through expert dialogues with ChatGPT$_{react}$, DUMA, and DUMA$_{StageI}$, each produced 80 groups of dialogues, named $test^{chatgpt}$, $test^{duma}$, $test^{dumaI}$. During the dialogue process, without affecting the dialogue logic, we reduce the deviation of the evaluation results by trying to ensure that the i-th group of test dialogue questions are the same:

\begin{align}
\begin{aligned}
& Q(test^{chatgpt}) \approx Q(test^{duma}) \\
& Q(test^{chatgpt}) \approx Q(test^{dumaI})
\end{aligned}
\end{align}

Each evaluation metric will be scored as 0, 1, or 2 points. The specific scoring criteria are detailed in the Table \ref{metric_cri}.

\subsection{Results and Analysis}

\begin{figure*}[hbt]
\vspace{0.1cm}
\centering
\setlength{\abovecaptionskip}{0.1cm} 
\includegraphics[scale=0.35]{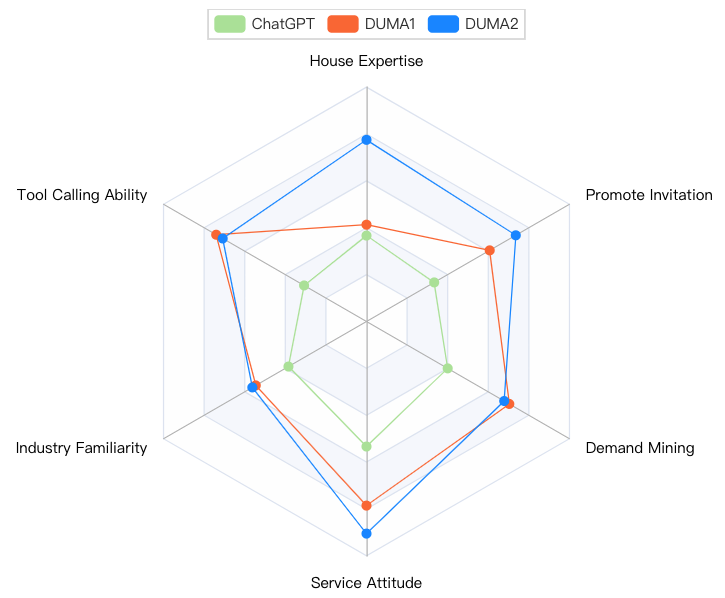}
\caption{Performance of ChatGPT, DUMA1, DUMA2 which respectively denotes ChatGPT$_{react}$, DUMA$_{StageI}$, DUMA.}
\label{duma_rador}
\end{figure*}

The experimental results are shown in Figure \ref{duma_rador}, with detailed scores presented in Table \ref{results}.

While ChatGPT produces relatively average results, it underperformes in every evaluation metric compared to both DUMA$_{StageI}$ and DUMA, demonstrating the effectiveness of the DUMA framework.

Through a two-stage SFT, DUMA displayes a noticeable improvement in \textit{House Expertise} over DUMA$_{StageI}$, gaining 0.724 points. Since in the second stage of SFT annotation, we corrected the erroneous appointment timings and service attitudes in the data, \textit{Promote Initiation} and \textit{Service Attitude} improved slightly, increasing by 0.257 points and 0.239 points respectively. These findings validate the effectiveness of the two-stage SFT.

Owing to the incorporation of 10\% of first stage dialogue data during the second stage SFT by $Mind_{Fast}$, DUMA's logical coherence remained relatively stable, with only minor decreases of 0.064 points in \textit{Tool Calling Ability} and 0.050 points in \textit{Demand Mining}. This demonstrates the viability of the mixed training approach in the second stage.

\section{Related work}

The development and potential of AI agents have been topics of significant interest in the AI community. An AI agent is defined as an artificial entity that senses its environment, makes decisions, and takes actions.\cite{zalta1995stanford,barandiaran2009defining}.

The emergence of Large Language Models (LLMs) is recognized as a potential catalyst for achieving Artificial General Intelligence (AGI) \cite{ouyang2022training,wei2022emergentabilities,bubeck2023sparksofAGI}. Recently, many works have proposed comprehensive LLM-based agent architectures\cite{weng2023llmpowered,wang2023survey,sumers2023cognitive,xi2023rise}. 
The key to dialogue agents being able to handle complex dialogue scenarios and apply knowledge lies in planning and tool utilization. 

\begin{itemize}
    \item \textbf{Planning}: LLMs exhibit Chain-of-Thought (CoT) reasoning, eliciting rationales through CoT prompts\cite{wei2022chain,wang2023selfconsistency,zhou2023least}. Yet, applying this reasoning in dialogues continues to be a challenge. ReAct\cite{yao2023react} defines a behavioral pattern of thinking and acting, allowing LLM to reason before each action planning.
    
    \item \textbf{Tool Use}: LLMs, as demonstrated by \cite{schick2023toolformer,li2023apibank,shen2023hugginggpt}, are adept at leveraging external resources, such as tools and APIs. The ability to extract knowledge from external sources has been showcased by works like WebGPT \cite{nakano2022webgpt} and ExpeL \cite{zhao2023expel}.
\end{itemize}

SwiftSage\cite{lin2023swiftsage} proposed an agent that combines fast and slow thinking, which is used in action planning for complex
interactive reasoning tasks. Our work also draws on the dual-process theory of human cognition, but we focus on building an Agent in a conversation scenario.

\section{Conclusions and Future work}

In this study, we introduced the DUMA framework, which intertwines the principles of fast and slow thinking within conversational scenarios. Our initial results, based on a specific Chinese real estate context, are promising. However, it's essential to approach these findings with caution until further validations in broader settings are conducted.

Our future efforts aim to test DUMA in more universal English-centric settings. Additionally, we recognize the need for a comparative study between standalone Slow Mind and Fast Mind versus their combined use. Future experiments will address these aspects, ensuring a clearer understanding and enhancing the framework’s versatility.

\bibliographystyle{acl}
\bibliography{acl2015}



\end{document}